\documentclass[letterpaper, 10 pt, conference]{ieeeconf}

\IEEEoverridecommandlockouts                              
\pdfoutput=1
\usepackage{cite}
\usepackage{url}

\usepackage[pdftex]{graphicx}
\usepackage{microtype}

\usepackage{amsmath} 
\usepackage{amssymb} 
\usepackage{siunitx}
\usepackage{breqn}

\usepackage[linesnumbered]{algorithm2e}
\usepackage{algorithmic}

\usepackage{array}

\usepackage{subfloat}
\usepackage{caption}
\usepackage[caption=false,font=footnotesize]{subfig}

\usepackage{pbox}
\usepackage{multirow}

\usepackage{color}

\newcommand{\vz}{{\bf z}}

\newcommand{\vZ}{{\bf Z}}
\newcommand{\vx}{{\bf x}}

\newcommand{\vS}{{\bf S}}
\newcommand{\vX}{{\bf X}}

\newsavebox{\ieeealgbox}

\title{A Multimodal Anomaly Detector for Robot-Assisted Feeding \\Using an LSTM-based Variational Autoencoder}

\author{Daehyung Park*, Yuuna Hoshi, and Charles C. Kemp
\thanks{D. Park, Y. Hoshi, and C. C. Kemp are with the Healthcare Robotics Lab, Institute for Robotics and Intelligent Machines, Georgia Institute of Technology. *D. Park is the corresponding author
\{\tt\small deric.park@gatech.edu\}.
}
} 

\begin{document}
\maketitle
\thispagestyle{empty}
\pagestyle{empty}

\begin{abstract}
The detection of anomalous executions is valuable for reducing potential hazards in assistive manipulation. Multimodal sensory signals can be helpful for detecting a wide range of anomalies. However, the fusion of high-dimensional and heterogeneous modalities is a challenging problem. We introduce a long short-term memory based variational autoencoder (LSTM-VAE) that fuses signals and reconstructs their expected distribution. We also introduce an LSTM-VAE-based detector using a reconstruction-based anomaly score and a state-based threshold. For evaluations with 1,555 robot-assisted feeding executions including 12 representative types of anomalies, our detector had a higher area under the receiver operating characteristic curve (AUC) of 0.8710 than 5 other baseline detectors from the literature. We also show the multimodal fusion through the LSTM-VAE is effective by comparing our detector with 17 raw sensory signals versus 4 hand-engineered features. 
\end{abstract}

\section{Introduction}
People with disabilities often need physical assistance from caregivers. Robots can provide assistance for activities of daily living such as robot-assisted feeding \cite{park2016towards} and shaving \cite{chen2013robots}. However, its structural complexity, task variability, and sensor uncertainty may result in failure. A lack of detection systems for the failures may also lower the usage of robots due to potential failure cost. The detection of an anomalous task execution (i.e., anomaly) can help to prevent or reduce potential hazards in the assistance by recognizing highly unusual situations and stop in these situations. 

In this paper, an anomaly detector is a method to identify when the current execution differs from past successful experiences (i.e., non-anomalous executions). Researchers often use an one-class classifier trained on non-anomalous execution. An ideal detector should detect a variety of anomalies, alert the robot quickly, ignore irrelevant task variation, and handle the stream of sensory signals. Multimodal sensory signals can be helpful to detect various anomalies using its high dimensional information. Researchers often reduce the dimension or select features before applying a classifier. Our previous work used also selected 4 hand-engineered features from 3 modalities for a likelihood-based classifier, HMM-GP, using hidden Markov models (HMM) \cite{park2016multimodal,park2017class}. However, the compressed or selected representations may be missing information relevant to anomaly detection. Creating useful hand-specified features can also involve significant engineering effort and domain expertise.

\begin{figure}[t]
	\centering
    \includegraphics[clip=true, width=6.8cm]{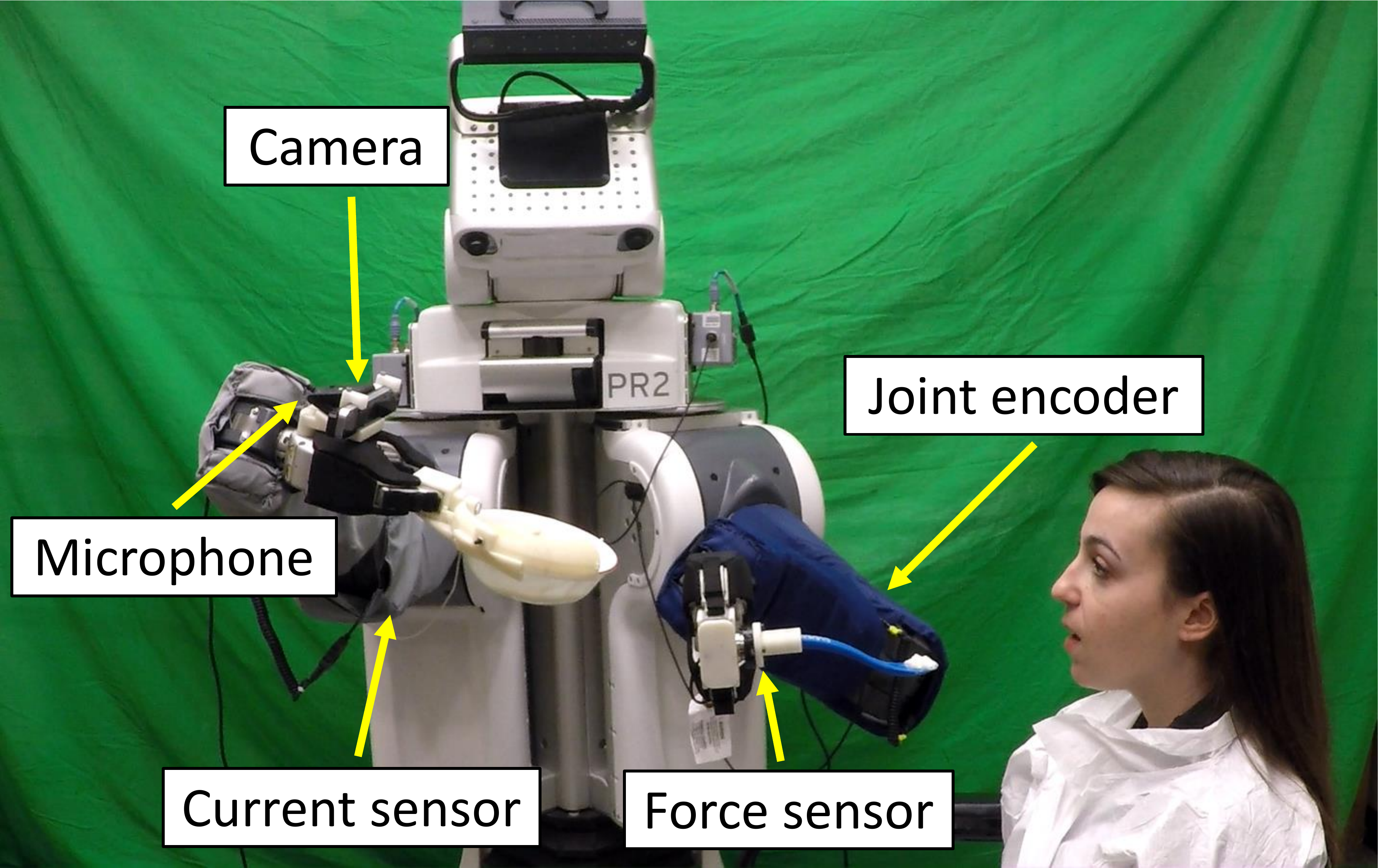}
	\caption{\textit{Robot-assisted feeding system. A PR2 robot detects anomalous feeding executions collecting 17 sensory signals from 5 types of sensors.}} 
    \vspace{-1.5em}
	\label{fig: sequence}
\end{figure}

An alternative solution is reconstruction-based detection, such as an autoencoder (AE) based approach that compresses and reconstructs high dimensional inputs based on non-anomalous executions. When an AE is trained only with non-anomalous data, a high reconstruction error can indicate an anomaly. The idea behind this detection is that an AE cannot reconstruct unforeseen patterns of anomalous data well compared to foreseen non-anomalous data. In addition to the reconstruction error, a variational autoencoder (VAE) can compute the reconstruction log-likelihood of the inputs modeling the underlying probability distribution of data. Both AE and VAE can be combined with time-series modeling approaches such as recurrent neural network (RNN) including long short-term memory (LSTM) network.

In this paper, we introduce a long short-term memory-based variational autoencoder (LSTM-VAE) for multimodal anomaly detection. For encoding, an LSTM-VAE projects multimodal observations and their temporal dependencies at each time step into a latent space using serially connected LSTM and VAE layers. For decoding, it estimates the expected distribution of the multimodal inputs from the latent space representation. We train it under a denoising autoencoding criterion \cite{vincent2008extracting} to prevent learning an identity function and improve representation capability. Our LSTM-VAE-based detector detects an anomaly when the log-likelihood of current observation given the expected distribution is lower than a threshold. We also introduce a state-based threshold to increase detection sensitivity and lower the false alarms similar to \cite{park2016multimodal}.

We evaluated the LSTM-VAE with robot-assisted feeding data that we collected from 24 able-bodied participants with 1,555 feeding executions. The proposed detector is beneficial in that we could directly use high-dimensional multimodal sensory signals without significant effort for feature engineering. It was able to catch an anomaly online. In particular, it was able to set tight or loose decision boundaries depending on the variations of multimodal signals using the state-based threshold. Our method had higher area under receiver operating characteristic (ROC) curves than other baseline methods from the literature. In our evaluation, the area under the curve (AUC) was 0.044 higher than that of our previous algorithm, HMM-GP given the same data. Our new method also had a 0.064 higher AUC when we used 17-dimensional sensory signals from visual, haptic, kinematic, and auditory modalities instead of 4-dimensional hand-engineered features.

\section{Related Work}
Anomaly detection is known as novelty, outlier, or event detections in other domains \cite{Chandola2009}. In robotics, it has been used to detect the failure of manipulation tasks: bin picking \cite{Rodriguez_2011_6893}, bottle opening \cite{Kappler-RSS-15}, etc. Many classic machine learning approaches have also been used: support vector machine (SVM) \cite{rodriguez2010failure, Hornung2014}, self-organizing map (SOM) \cite{haussermann2014novel}, k-nearest neighbors (kNN) \cite{ando2011ace}, etc. To detect anomalies from time-series signals, researchers have also used hidden Markov models \cite{park2016multimodal} or Kalman filters \cite{furukawa2015estimating}.

Researchers have often reduced the dimension of high-dimensional inputs using principal component analysis (PCA) before applying probabilistic or distance-based detection \cite{rodriguez2010failure, sukhoy2012learning}. However, the compressed representations of outliers (i.e., anomalous data) may be inliers in latent space. Instead, we use a reconstruction-based method that recovers inputs from its compressed representation so that it can measure reconstruction error with the anomaly score. An AE is a representative reconstruction approach that is a connected network with an encoder and a decoder \cite{hinton2006reducing}. It has also been applied for reconstructing time-series data using a sliding time-window \cite{noda2014multimodal}. However, the window method does not represent dependencies between nearby windows and a window may not include an anomaly.

  To model time-series data with its temporal dependencies, we use an LSTM network \cite{hochreiter1997long}, which is a type of recurrent neural network (RNN). An LSTM network can make use of long-term dependencies and avoid the vanishing gradient problem \cite{hochreiter1997long}. Researchers have used LSTM networks for prediction in this anomaly detection domains such as the following: radio anomaly detection \cite{o2016recurrent} and EEG signal anomaly detection\cite{chauhan2015anomaly}. Malhorta et al. introduced an LSTM-based anomaly detector (LSTM-AD) that measures the distribution of prediction errors \cite{malhotra2015long}. However, the method may not predict time-series under unpredictable external changes such as manual control and load on a machine \cite{malhotra2016lstm}. Alternatively, researchers have introduced RNN- and LSTM-based autoencoders for reconstruction-based anomaly detection \cite{cho2014learning, principi2017acoustic}. In particular, Malhorta et al. introduced an LSTM-based encoder-and-decoder (EncDec-AD) that estimates a reconstruction error \cite{malhotra2016lstm}. We also use this reconstruction scheme for detection and as a baseline method in this paper.

Another relevant approach is a variational autoencoder (VAE) \cite{kingma2013auto}. Unlike the AE, a VAE models the underlying probability distribution of observations using variational inference (VI). Bayer and Osendorfer used VI to learn the underlying distribution of sequences and introduced stochastic recurrent networks \cite{bayer2014learning}. Soelch et al. used their work to detect robot anomalies using unimodal signals \cite{solch2016variational}. Our work also uses variational inference, but we do not predict and instead only reconstruct data using an LSTM-based autoencoder for multimodal anomaly detection.

\section{LSTM-based Variational Autoencoding}
We review an autoencoder and a variational autoencoder. We then describe our proposed LSTM-based variational autoencoder. We represent a vector of multidimensional inputs by $\vx \in \mathbb{R}^{D}$ and the corresponding latent space vector by $\vz \in \mathcal{R}^K$, where $D$ and $K$ are the number of input signals and the dimension of the latent space, respectively. 

\subsection{Preliminary: Autoencoder(AE)}
An AE is an artificial neural network that consists of sequentially connected encoder and decoder networks. It sets the target of the decoder to be equal to the input of the encoder. The encoder network learns a compressed representation (i.e., bottleneck feature or latent variable) of the input. The decoder network reconstructs the target from the compressed representation. The difference between the input and the reconstructed input is the reconstruction error. During training, the autoencoder minimizes the reconstruction error as an objective function. An AE is often used for data generation as a generative model. An AE's decoder can generate an output given an artificially assigned compressed representation.

\begin{figure*}[t]
	\centering
    \vspace{8pt}
	\includegraphics[trim=0.0cm 0.0cm 0.0cm 0.0cm, clip=true, width=17.0cm]{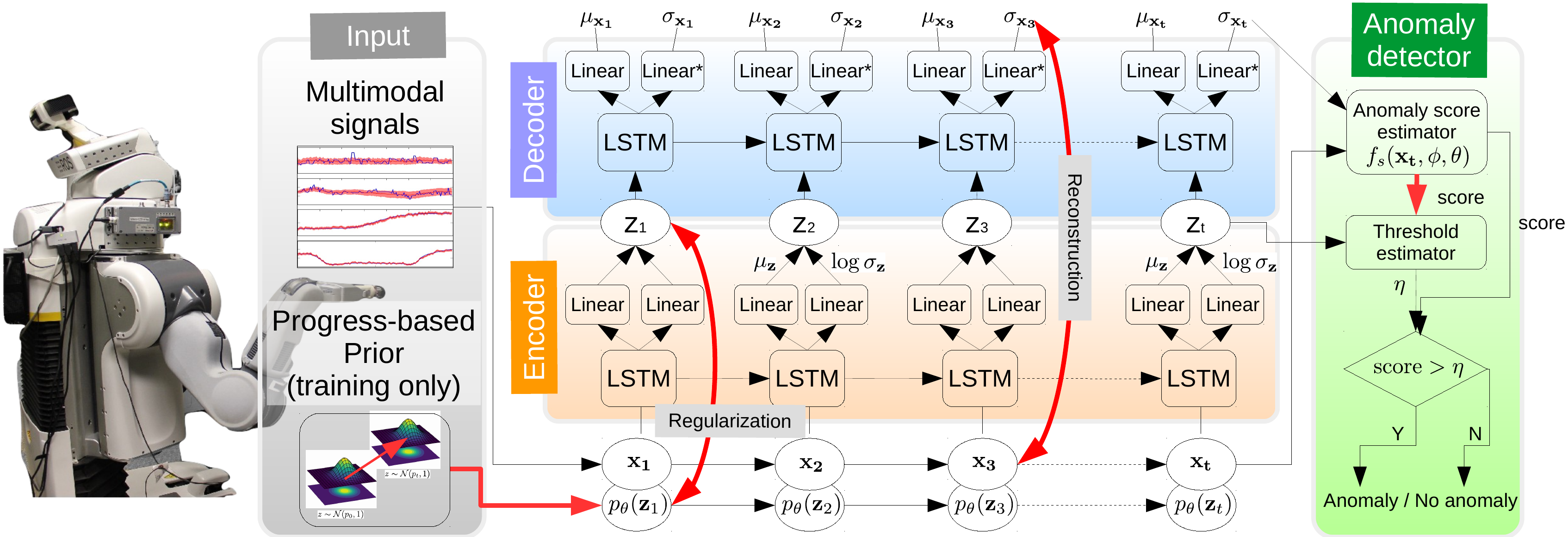} 	
	\caption{\textit{Illustration of a multimodal anomaly detector with an unrolled LSTM-VAE model. We train the LSTM-VAE using multimodal signals and corresponding progress-based priors. We then train a threshold estimator using the outputs of the LSTM-VAE. For testing, we input sensory signals only. The detector then returns an anomaly when current anomaly score is over an estimated threshold $\eta$. Note that Linear* and LSTM layers have tanh and softplus activations, respectively. The red arrows are used for training only.}}
	\label{fig: structure}
    \vspace{-1.0em}
\end{figure*}

\subsection{Preliminary: Variational Autoencoder(VAE)}
A VAE is a variant of an AE rooted in Bayesian inference \cite{kingma2013auto}. A VAE is able to model the underlying distribution of observations $p(\vz)$ and generate new data by introducing a set of latent random variables $\vz$. We can represent the process as $p(\vx)=\int p(\vx|\vz)p(\vz) d\vz$. However, the integral is intractable due to the continuous domain of $\vz$. Instead, we can represent the marginal log-likelihood of an individual point as $\log p(\vx) = D_{KL}(q_{\phi}(\vz|\vx)||p_{\theta}(\vz)) +\mathcal{L_{\mathrm{vae}}}(\phi, \theta;\vx)$ using notation from \cite{kingma2013auto}, where $D_{KL}$ is Kullback\-Leibler divergence from a prior $p_{\theta}(\vz)$ to the variational approximation $q_{\phi}(\vz|\vx)$ of $p(\vz|\vx)$ and $\mathcal{L_{\mathrm{vae}}}$ is the variational lower bound of the data $\vx$ by Jensen's inequality. Note that $\phi$ and $\theta$ are the parameters of the encoder and the decoder, respectively.

A VAE optimizes the parameters, $\phi$ and $\theta$, by maximizing the lower bound of the log likelihood, $\mathcal{L_{\mathrm{vae}}}$,
\begin{align}
\mathcal{L_{\mathrm{vae}}} &= -D_{KL}(q_{\phi}(\vz|\vx)||p_{\theta}(\vz)) + \mathbb{E}_{q_{\phi}(\vz|\vx)}[\log p_{\theta}(\vx|\vz)].
\label{eq_lower_bound}
\end{align}%
The first term regularizes the latent variable $\vz$ by minimizing the KL divergence between the approximated posterior and the prior of the latent variable. The second term is the reconstruction of $\vx$ by maximizing the log-likelihood $\log p_{\theta}(\vx|\vz)$ with sampling from $q_{\phi}(\vz|\vx)$. 

The choice of distribution types are important since a VAE models the approximated posterior distribution $q_\phi(\vz|\vx)$ from a prior $p_{\theta}(\vz)$ and likelihood $p_{\theta}(\vx | \vz)$. A typical choice for the posterior is a Gaussian distribution, $\mathcal{N}(\mu_{\vz},\Sigma_{\vz})$, where a standard normal distribution $\mathcal{N}(0,1)$ is used for the prior. For the likelihood, a Bernoulli distribution or multivariate Gaussian distribution is often used for binary or continuous data, respectively.

\subsection{An LSTM-based Variational Autoencoder (LSTM-VAE)} \label{ss_lstm_vae}
We introduce a long short-term memory-based variational autoencoder (LSTM-VAE). To use the temporal dependency of time-series data in a VAE, we combine a VAE with LSTMs by replacing the feed-forward network in a VAE to LSTMs similar to conventional temporal AEs such as an RNN Encoder-Decoder \cite{cho2014learning} or an EncDec-AD \cite{malhotra2016lstm}. Fig. \ref{fig: structure} shows an unrolled structure with LSTM-based encoder-and-decoder modules. Given a multimodal input $\vx_t$ at time $t$, the encoder approximates the posterior $p(\vz_{t}|\vx_{t})$ by feeding an LSTM's output into two linear modules to estimate the mean $\mu_{\vz_{t}}$ and co-variance $\Sigma_{\vz_{t}}$ of the latent variable. Then, the randomly sampled $\vz$ from the posterior $p(\vz_{t}|\vx_{t})$ feeds into the decoder's LSTM. The final outputs are the reconstruction mean $\mu_{\vx_{t}}$ and co-variance $\Sigma_{\vx_{t}}$. 

We apply a denoising autoencoding criterion \cite{vincent2008extracting} to the LSTM-VAE by introducing corrupted input with Gaussian noise, $\tilde{\vx} = \vx + \epsilon$, where $\epsilon \sim \mathcal{N}(0, \sigma_{\mathrm{noise}})$. We then replace the lower bound in Eq. (\ref{eq_lower_bound}) with a denoising variational lower bound $\mathcal{L_{\mathrm{dvae}}}$ \cite{im2017denoising}, 
\begin{align}
\mathcal{L_{\mathrm{dvae}}} &= -D_{KL}(\tilde{q}_{\phi}(\vz_{t}|\vx_{t})||p_{\theta}(\vz_{t})) \nonumber \\
&+\mathbb{E}_{\tilde{q}_{\phi}(\vz_t|\vx_t)}[\log p_{\theta}(\vx_t|\vz_t)],
\label{eq_denoising_lower_bound}
\end{align}%
where $\tilde{q}_{\phi}(\vz_t | \vx_t )$ is an approximated posterior distribution given a corruption distribution around $\vx_t$. Given Gaussian distribution for $p(\tilde{\vx}|\vx)$ and $q_{\phi}(\vz|\vx)$, $\tilde{q}_{\phi}(\vz_t | \vx_t )$ can be represented as a mixture of Gaussians. For computational convenience, we use a single Gaussian, $\tilde{q}_{\phi}(\vz|\vx) \approx q_{\phi}(\vz|\tilde{\vx})$.

\begin{figure}[t]
	\centering
	\includegraphics[trim=0.0cm 0.0cm 0.0cm 0.0cm, clip=true, width=7.5cm]{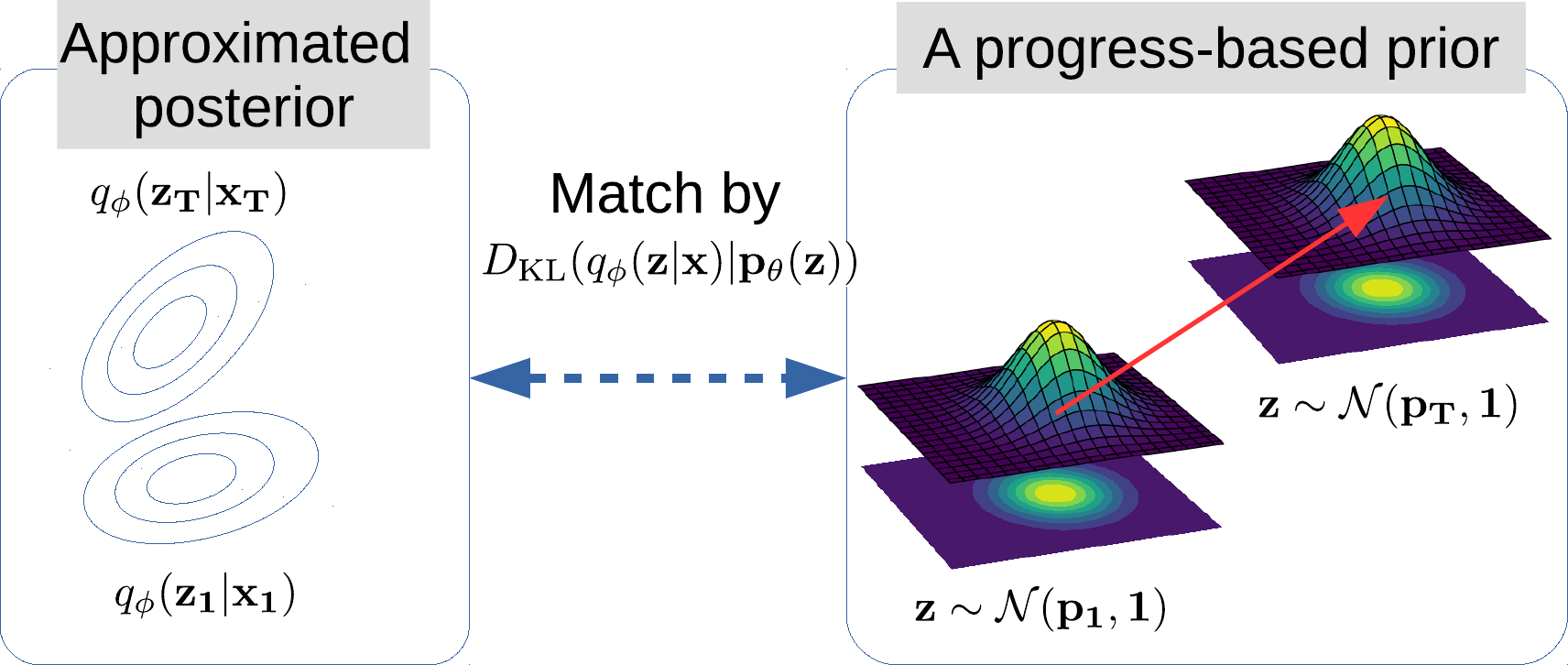} 
	\caption{\textit{Illustration of the progress-based prior. The center of the prior linearly changes from $p_1$ as initial progress to $p_T$ as final progress.}}
	\label{fig: progress_prior}
    \vspace{-1.5em}
\end{figure}

We introduce a progress-based prior $p(\vz_{t})$. Unlike conventional static priors using a normal distribution $\mathcal{N}(0,1)$, we vary the center of a normal distribution as $\mathcal{N}(\mu_p,\Sigma_p)$, where $\mu_p$ and $\Sigma_p$ are the center and co-variance of the underlying distribution of multimodal inputs, respectively (see Fig. \ref{fig: progress_prior}). This kind of varying prior can be helpful to introduce the temporal dependency of time-series data into its underlying distribution since a VAE tries to minimize the difference between the approximated posterior and the prior. Unlike the RNN prior of Solch et al. \cite{solch2016variational} and the transition prior of Karl et al. \cite{karl2016deep}, we gradually change $\mu_p$ from $p_1$ to $p_T$ as the progress of a task execution. To simplify the prior, we use an isotropic normal distribution so the co-variance matrix is $\Sigma_p=I$. We can rewrite the regularization term of $\mathcal{L_{\mathrm{dvae}}}$ as
\begin{align}
&D_{KL}(\tilde{q}_{\phi}(\vz_{t}|\vx_{t})||p_{\theta}(\vz_{t})) \nonumber \\
&\approx D_{KL}(\mathcal{N}(\mu_{\vz_{t}},\Sigma_{\vz_{t}})||\mathcal{N}(\mu_p,1)). \nonumber \\
&={ \frac{1}{2} } \left( \mathrm{tr}( \Sigma_{\vz_{t}} ) + ( \mu_p - \mu_{\vz_{t}})^T ( \mu_p - \mu_{\vz_{t}} ) 
 - D - \log { | \Sigma_{\vz_{t}} |  } \right).
\end{align}

To represent the distribution of high-dimensional continuous data, we use a multivariate Gaussian with a diagonal co-variance matrix. We can derive the reconstruction term in $\mathcal{L_{\mathrm{dvae}}}$ as
\begin{align}
&\mathbb{E}_{\tilde{q}_{\phi}(\vz_t|\vx_t)}[\log p_{\theta}(\vx_t|\vz_t)] \nonumber \\
&= -\frac{1}{2} ( \log(|\Sigma_{\vx_{t}} |) + (\vx_{t} - \mu_{\vx_{t}})^T \Sigma_{\vx_{t}}^{-1} ({\vx_{t}} - \mu_{\vx_{t}}) \nonumber \\
&+ D \log(2\pi) ) 
\label{eq_lower_bound_new}
\end{align}

We use an LSTM with \textit{tanh} for each encoder and decoder. We implemented the LSTM-VAE using stateful LSTM models in the Keras deep learning library \cite{chollet2015keras}. We trained the LSTM-VAE using an Adam optimizer with 3-dimensional latent variables and a 0.001 learning rate. Note that we are not using a sliding window in this work, but a window could be applied.

\section{Anomaly Detection}
We now introduce an online anomaly detection framework for multimodal sensory signals with state-based thresholding. 

\subsection{Anomaly Score}
Our method detects an anomalous execution when the current anomaly score of an observation $\vx_t$ is higher than a score threshold $\eta$, 
\begin{align} 
\begin{cases}
\mathrm{anomaly}, & \text{if } f_s(\vx_t, \phi, \theta) > \eta \\
\lnot\mathrm{anomaly}, & \text{otherwise,}
\end{cases}
\label{eq_stc_threshold}
\end{align}
where $f_s(\vx_t, \phi, \theta)$ is an anomaly score estimator. We define the score as the negative log-likelihood of an observation with respect to the reconstructed distribution of the observation through an encoding-decoding model, 
\begin{align}
f_s(\vx_{t}, \phi, \theta) &= -\log p(\vx_t ; \mu_{\vx_t}, \Sigma_{\vx_t}), 
\label{eq_anomaly_score}
\end{align}
where $\mu_{\vx_t}$ and $\Sigma_{\vx_t}$ are the mean and co-variance of the reconstructed distribution, $\mathcal{N}(\mu_{\vx_t},\Sigma_{\vx_t})$, from an LSTM-VAE with parameters $\phi$ and $\theta$. A high score indicates an input has not been reconstructed well by the the LSTM-VAE. In other words, the input has deviated greatly from the non-anomalous training data.

\subsection{State-based Thresholding}
We introduce a varying threshold that changes over the estimated state of a task execution motivated by the dynamic threshold \cite{park2016multimodal}. Depending on the state of task executions, reconstruction quality may vary. In other words, anomaly scores in non-anomalous task executions can be high in certain states, so varying the anomaly score can reduce false alarms and improve sensitivity. In this paper, the state is the latent space representation of observations. Given a sequence of observations, the encoder of LSTM-VAE is able to compute a state at each time step. By mapping states $\vZ$ and corresponding anomaly scores $\vS$ from non-anomalous dataset, our method is able to train an expected anomaly score estimator $\hat{f}_s: \vz \rightarrow s$. We use support vector regression (SVR) to map from a multidimensional input $\vz \in \vZ$ to a scaler $s$ using a radial basis function (RBF) kernel. To control sensitivity, we add a constant $c$ into the expected score and represent the state-based threshold as $\eta = \hat{f}_s(\vz) + c$.

\subsection{Training and Testing Framework}
Algorithm \ref{alg: training} shows the training framework of our LSTM-VAE-based anomaly detector. Given a set of non-anomalous training and validation data, $(\vX_{\mathrm{train}}, \vX_{\mathrm{val}})$, the framework aims to output the optimized parameters $(\phi, \theta)$ of an LSTM-VAE and an expected anomaly score estimator $\hat{f}_s$. Note that we represent $N$ sequences of multimodal observations as $\vX = \{\vx^{(1)}, \vx^{(2)}, ... , \vx^{(N)} \}$. $N_{\mathrm{train}}$ and $N_{\mathrm{val}}$ denote the numbers of training and validation data, respectively. We also represent the encoder and decoder functions as $f_\phi: \vx_{t} \rightarrow \vz_{t}$ and $g_\theta: \vz_{t} \rightarrow (\mu_{\vx_t}, \Sigma_{\vx_t})$, respectively. Then, we denote the function of serially connected encoder and decoder (i.e., autoencoder) by $f_{\phi, \theta}$ with noise injection.

The framework preprocesses $\vX_{\mathrm{train}}$ and $\vX_{\mathrm{val}}$ by resampling those to have length $T$ and normalizing their individual modalities in the range of $[0,1]$ with respect to $\vX_{\mathrm{train}}$. The framework then starts to train the LSTM-VAE with respect to $\vX_{\mathrm{train}}$ maximizing $L_{\mathrm{dvae}}$ and stops the training when $L_{\mathrm{dvae}}$ does not increase for 4 epochs. Then it extracts a set of latent space representations and corresponding anomaly scores from $\vX_{\mathrm{val}}$ as the training set for $\hat{f}_s$. Finally, this framework returns the trained SVR object as well as the LSTM-VAE's parameters. Note that we reset the state of the LSTM in the beginning of a sequence of data only. 

In testing, the detector aims to detect an anomaly in real time. Algorithm \ref{alg: testing} shows the pseudo code for the online detection process. In each loop, the detector takes multimodal input \vx. The detector scales individual dimension with respect to the scaled $\vX_{\mathrm{train}}$. It then estimates the latent variable and the parameters of the reconstructed distribution. When the anomaly score of the current input is higher than the threshold $\eta$, our system detects that the detector determines the current task execution is anomalous and returns the decision. We control the sensitivity of the detector by adding a constant $c$ to $\hat{f}_s$.

\RestyleAlgo{boxruled}
\begin{algorithm}
 \SetKwInOut{Input}{input}\SetKwInOut{Output}{output}\SetKw{Return}{return}
 \Input{$\vX_{\mathrm{train}} \in \mathbb{R}^{N_{\mathrm{train}} \times T \times D}$, $\vX_{\mathrm{val}} \in \mathbb{R}^{N_{\mathrm{val}} \times T \times D}$}
 \Output{$\phi$, $\theta$, $f_\eta$}
 \BlankLine
$\vX_{\mathrm{train}}, \vX_{\mathrm{val}} $ = Preprocessing($\vX_{\mathrm{train}}, \vX_{\mathrm{val}}$) \;
$\phi ,\theta \leftarrow$ train LSTM-VAE with ($\vX_{\mathrm{train}},\vX_{\mathrm{val}})$\;
$\vZ = \emptyset, \vS = \emptyset$ \;
\For{$i\gets1$ \KwTo $N_{\mathrm{val}}$}{
	Reset the state of LSTM-VAE\;
	\For{$j\gets1$ \KwTo $T$}{
		$\vz \leftarrow f_{\phi}(\vX_{\mathrm{val}}(i,j))$\;
        $\mu_x, \sigma_\vx \leftarrow f_{\phi, \theta}(\vX_{\mathrm{val}}(i,j))$\;
		$s \leftarrow f_{s}(\vx_{\mathrm{val}}(i,j), \mu_\vx, \sigma_\vx)$\;
        Add $\vz$ and $s$ into $\vZ$ and $\vS$, respectively.
    }	
} 
$\hat{f}_{s} \leftarrow$ train an SVR with $(\vZ, \vS)$.
 \caption{Training algorithm for an LSTM-VAE-based anomaly detector}\label{alg: training}
\end{algorithm}

\RestyleAlgo{boxruled}
\begin{algorithm}
 \SetKwInOut{Input}{input}\SetKwInOut{Output}{output}\SetKw{Return}{return}
 \Input{$\vx \in \mathbb{R}^{D}$}
 \Output{Anomaly or $\neg$Anomaly}
 \BlankLine
 \While{True}{
 	$\vx \leftarrow$ get current multimodal data\;
    $\vx \leftarrow Preprocessing(\vx)$\;
	$\vz \leftarrow f_{\phi}(\vx)$\;
 	$\mu_\vx, \Sigma_\vx \leftarrow f_{\phi, \theta}(\vx)$\;
 	\If{$f_{s}(\vx ;\mu_\vx, \Sigma_\vx ) > \hat{f}_s(\vz)+c$}{
 		\Return Anomaly \;
 	}
 }
 \caption{Testing algorithm for an LSTM-VAE-based anomaly detector.}\label{alg: testing}
\end{algorithm}

\section{Experimental Setup}

\subsection{Instrumental Setup}
Our system uses a PR2 from Willow Garage, a general-purpose mobile manipulator with two 7-DOF arms with powered grippers and an omni-directional mobile base. For safety and prevention of possible hazards, we used a low-level PID controller with low gains and a \SI{50}{\hertz} mid-level model predictive controller from~\cite{jain2013reaching} without haptic feedback. We used the following sensors: an RGB-D camera with a microphone (Intel SR300) on the right wrist, a force/torque sensor (ATI Nano25) on the utensil handle, joint encoders, and current sensors. These sensors measure mouth position and sound, force on the utensil, spoon position, and joint torque, respectively. 

\subsection{Data Collection}
We used data from 1,555 feeding executions collected from 24 able-bodied participants where we newly collected 1,203 non-anomalous feeding executions for this work. 16 participants were male and 8 were female, and the age range was 19-35. We conducted the studies with approval from the Georgia Tech Institutional Review Board (IRB).

We divided our data into two subsets: a training/testing dataset collected from our previous work \cite{park2017class}, and a pre-training dataset. The training/testing dataset consists of data from 352 executions (160 anomalous and 192 non-anomalous) collected from 8 able-bodied participants who used the feeding system with yogurt and a silicone spoon. 

The pre-training dataset uses data from 1,203 non-anomalous executions from 16 newly recruited participants who used various foods and utensils. Pre-training with this dataset allowed us to initialize the weights of the LSTM-VAE to find a better fit in fine tuning. Among the dataset, 559 non-anomalous executions were from 9 participants who used 3 types of food and corresponding utensils: cottage cheese and silicone spoon, watermelon chunks and metal fork, and fruit mix and plastic spoon. An experimenter also conducted 428 non-anomalous feeding executions as a self-study with 6 foods (yogurt, rice, fruit mix, watermelon chunks, cereal, and cottage cheese) and 5 utensils (small/large plastic spoons, a silicone spoon, and plastic/metal forks). We also collected additional data from 216 non-anomalous executions from 6 participants who used yogurt and a silicone spoon.

\subsection{Experimental Procedure}
This feeding system allows the user to command three autonomous subtasks: \textit{scooping/stabbing}, \textit{clean spoon}, and \textit{feeding}. The user sends these commands using a web-based graphical user interface. A typical feeding sequence consists of scooping or stabbing followed by feeding. In order to approximate one form of limited mobility that people with disabilities might have, we instructed the participants to not move their upper bodies and to eat food off of the utensil using their lips.

\begin{figure}
	\centering
    \vspace{8pt}
    \includegraphics[trim=0.5cm 0.0cm 0.7cm 0.0cm, clip=true, width=8.0cm]{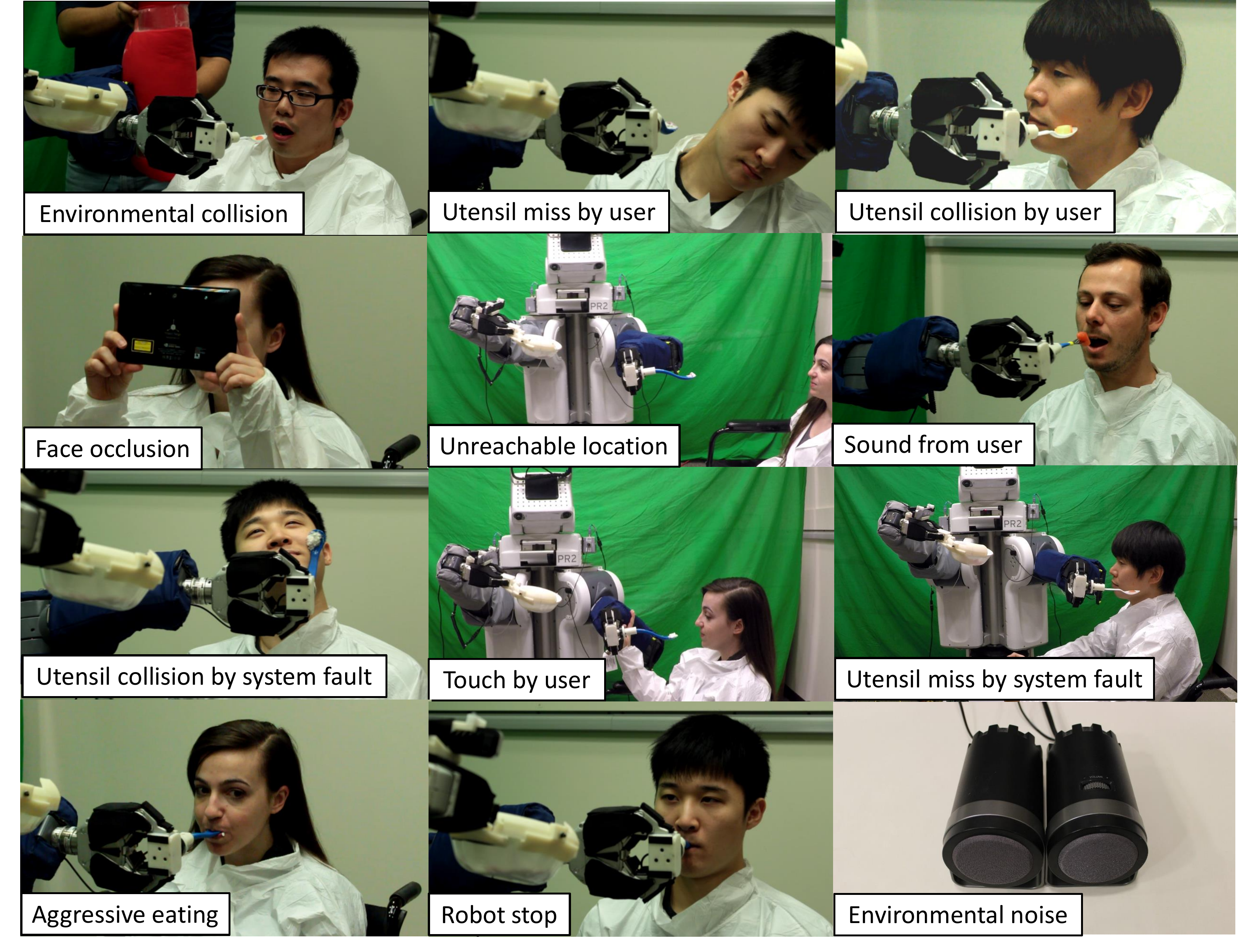}
    \caption{\textit{12 representative anomalies caused by either the user, the environment, or the system in our experiments.}}
    \label{fig: anomaly}
    \vspace{-1.5em}    
\end{figure}

Each participant performed anomalous and non-anomalous feeding executions while the participant, experimenters, or the system produced anomalies. We randomly determined the order of these executions. We defined 12 types of representative anomalies through fault tree analysis \cite{ogorodnikova2008methodology}: touch by user, aggressive eating, utensil collision by user, sound from user, face occlusion, utensil miss by user, unreachable location, environmental collision, environmental noise, utensil miss by system fault, utensil collision by system fault, and system freeze (see Fig. \ref{fig: anomaly}). For anomalies caused by the user, we instructed the participants on their actions through demonstration videos and verbal explanation. The participant had control over the details of the anomaly, such as the exact timing and magnitude of their actions.

\begin{figure}
	\centering
    \vspace{8pt}
    \includegraphics[width=7.0cm, clip=true]{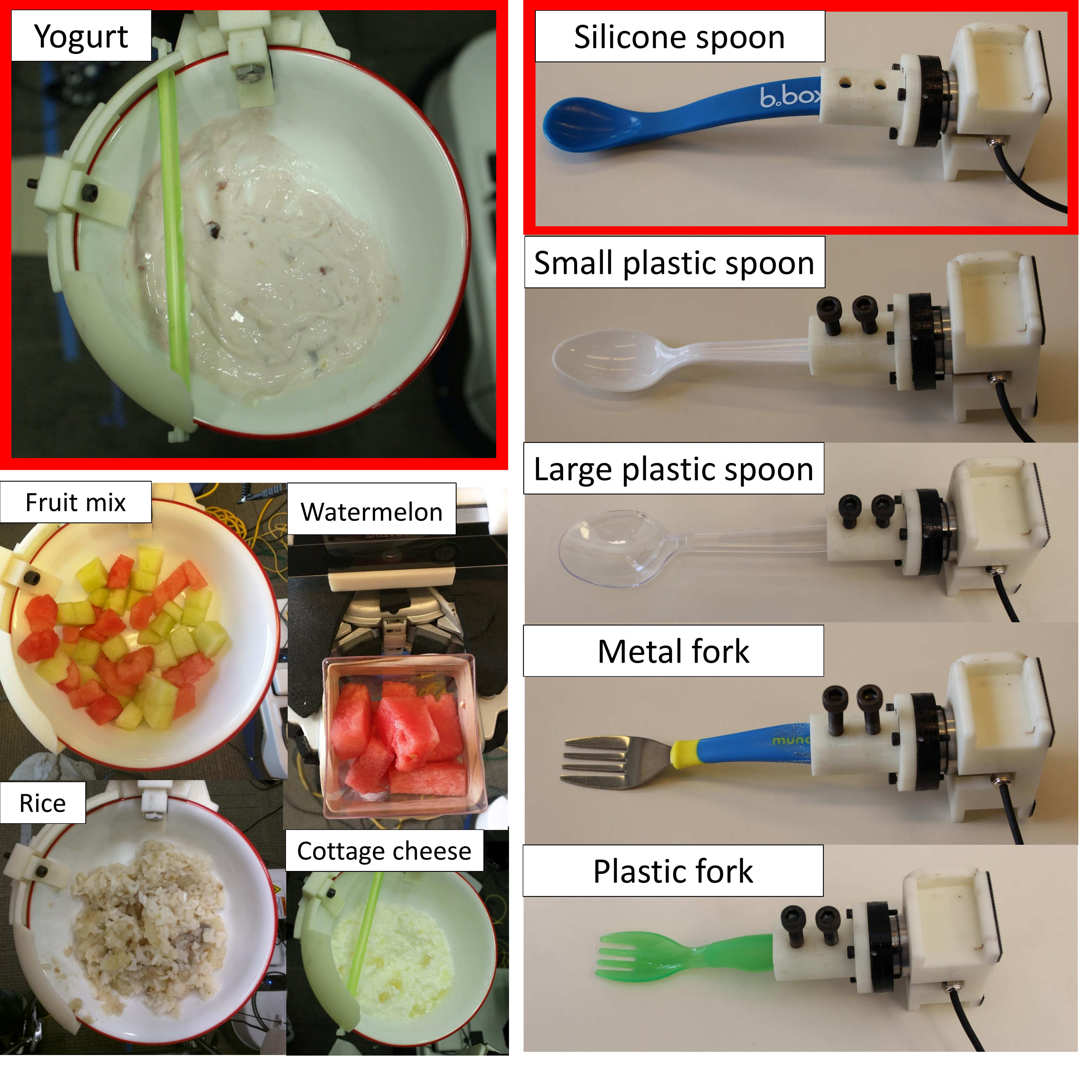}
    \caption{\textit{\textbf{Left:} Examples of food used in our experiments. \textbf{Right:} The 3D-printed utensil handle and 5 utensils used. Red boxes show yogurt and silicone spoon used for our training/testing dataset.}}
    \label{fig: feeding_tool}
    \vspace{-1.5em}    
\end{figure}

\subsection{Data collection and Pre-processing} \label{ss_data_collection}
For each feeding execution, we collected 17 sensory signals from 5 sensors: 
\textit{sound energy} (1), \textit{force} (3) applied on the end effector, \textit{joint torque} (7), \textit{spoon position} (3), and \textit{mouth position} (3), where the number in parentheses represents the dimension of signals. We zeroed the initial value and resampled each signal to have \SI{20}{Hz} for the robot's actual anomaly check frequency. We then scale signals in the non-anomalous dataset to have a value between 0 and 1. Corresponding to this scale, we also scaled signals from the anomalous dataset. Finally, we have a sequence of tuples per execution (i.e., \textit{sequence length} $\times$ 17). For visualization and comparison purposes, we also extracted 4-dimensional hand-engineered features used in our previous work \cite{park2017detection}: \textit{sound energy}, \textit{1st joint torque}, \textit{accumulated force}, and \textit{spoon-mouth distance}. Here, we used \textit{sound energy}\footnote{Root mean square (RMS) of 1,024 frames.} instead of raw \SI{44100}{\kHz} 16 bit PCM encoding since the under sampling could miss auditory anomalies.

\begin{figure}
	\centering
    \subfloat[A non-anomalous execution]{
		\includegraphics[trim=1.0cm 0.4cm 1.0cm 0.7cm, clip=true, width=8cm]{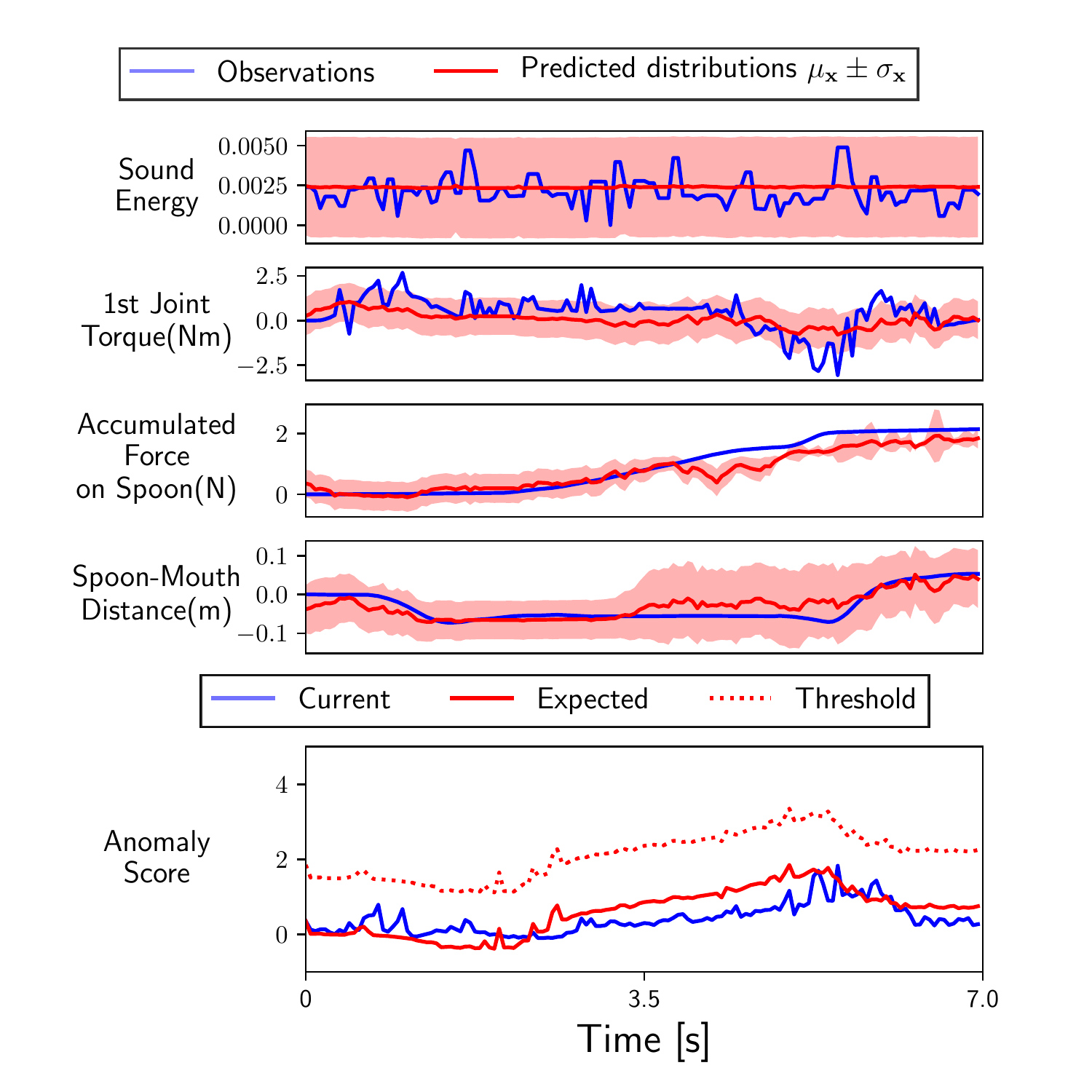} 
        \label{fig: reconstruction_normal}
    }\\
    \subfloat[An anomalous execution]{
	\includegraphics[trim=1.0cm 0.4cm 1.0cm 1.8cm, clip=true, width=8cm]{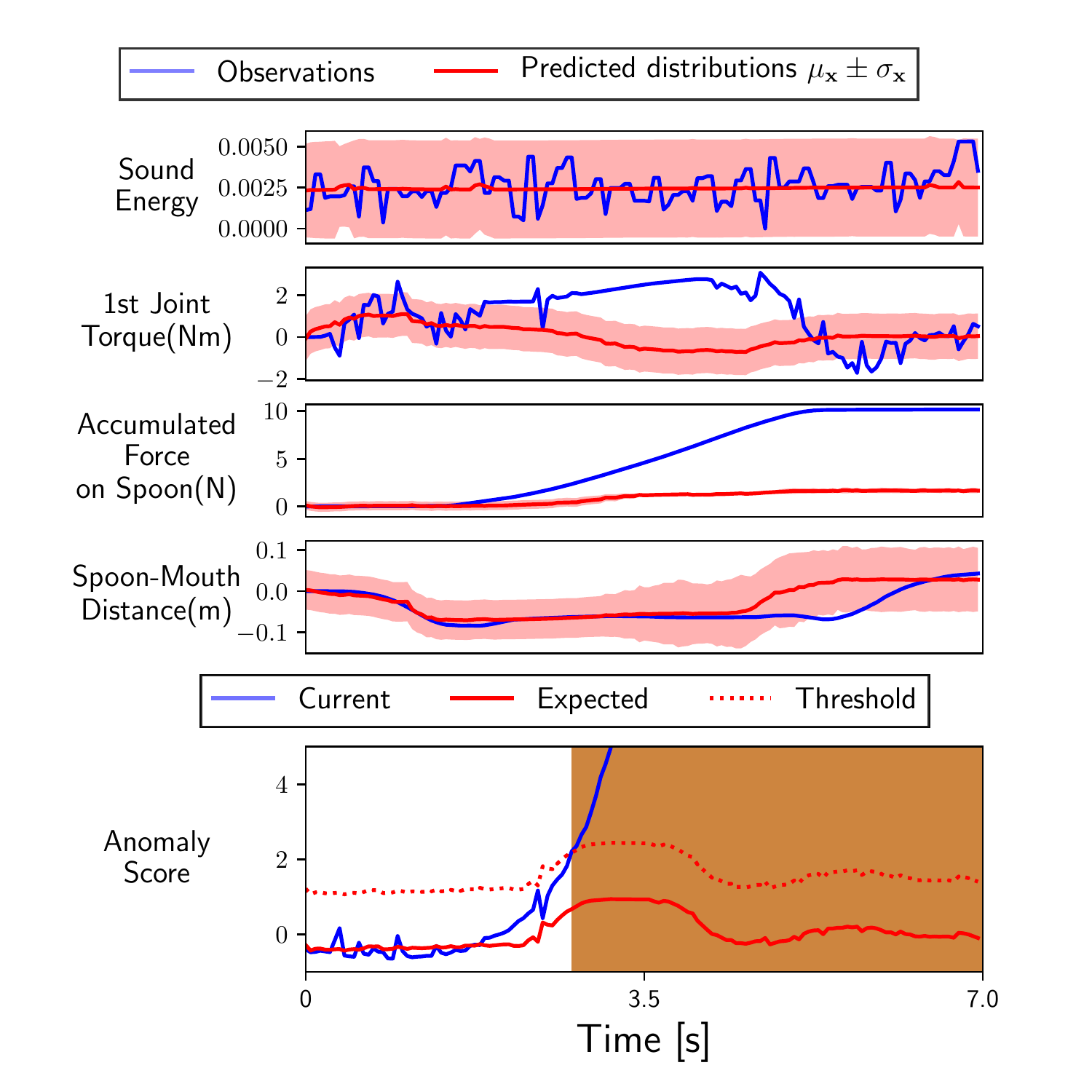} 
    	\label{fig: reconstruction_abnormal}
    }
	\caption{\textit{Visualization of the reconstruction performance and anomaly scores over time using an LSTM-VAE. The upper four sub graphs show observations and reconstructed observations' distribution. The lower sub graphs show current and expected anomaly scores. The dashed curve shows a state-based threshold where the LSTM-VAE determines an anomaly when current anomaly score is over the threshold. Brown lines represent the time of anomaly detection.}}
    \vspace{-1.5em}
	\label{fig: reconstruction}
\end{figure}

\subsection{Baseline Methods}
To evaluate the performance of the proposed method, we implemented 5-baseline methods,
\begin{itemize}
\item RANDOM: A random binary classifier in which we control its sensitivity by weighting a class.
\item OSVM: A one-class SVM-based detector trained with only non-anomalous executions. We move a sliding window (of size 3 in time like EncDec-AD \cite{malhotra2016lstm}) one step at a time. We control its sensitivity by adjusting the number of support vectors.
\item HMM-GP: A likelihood-based classifier using an HMM introduced in \cite{park2017detection}. We vary the likelihood threshold with respect to the distribution of hidden states.
\item AE: A reconstruction-based anomaly detector using a conventional autoencoder with a 3 time step sliding window based on \cite{an2015variational}.
\item EncDec-AD: A reconstruction-based anomaly detector using an LSTM-based autoencoder \cite{malhotra2016lstm}. We use window size $L=3$ as in the paper, but unlike the paper we use a diagonal co-variance matrix when we model the distribution of reconstruction-error vectors.
\end{itemize}
From now on, we will also use the term LSTM-VAE to refer to our LSTM-VAE-based detector.

\section{Evaluation}

We first investigated the reconstruction function of the LSTM-VAE. The upper 4 sub graphs in Fig. \ref{fig: reconstruction} show the reconstructed distribution of 4 hand-engineered features from non-anomalous and anomalous feeding executions in the robot-assisted feeding task. For Fig. \ref{fig: reconstruction_normal}, the observed features (blue curves) and the mean of reconstructed distribution (red curves) show a similar pattern of change over time. On the other hand, in an anomalous execution (see Fig. \ref{fig: reconstruction_abnormal}), the LSTM-VAE resulted in a large deviation between observed and reconstructed \textit{accumulated force} since the pattern of accumulated force by the collision is not easily observable from non-anomalous executions. Consequently, we can observe the anomaly score (blue curve) gradually increases after the onset of the deviation from the lower sub graph of Fig. \ref{fig: reconstruction_abnormal}. Note that the anomalous execution came from a face-spoon collision caused intentionally by the user.

\begin{figure}[t]
	\centering
	\includegraphics[trim=0.0cm 0.0cm 0.0cm 0.0cm, clip=true, width=7.0cm]{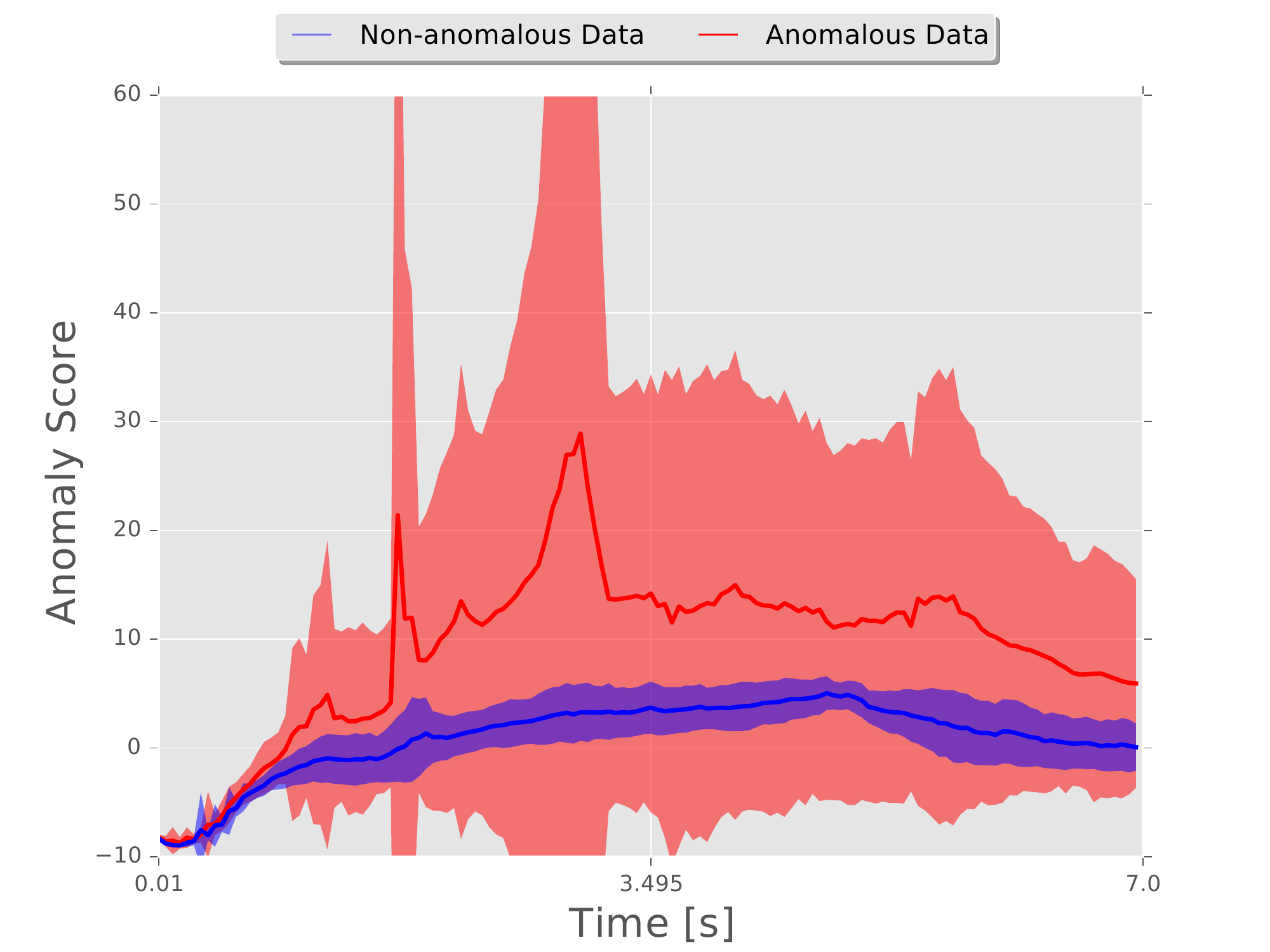} 
	\caption{\textit{An example distributions of anomaly scores from a participant's 20 non-anomalous and 24 anomalous executions over time.}}
    \vspace{-1.5em}
	\label{fig: anomaly_score}
\end{figure}

The anomaly score metric is effective in distinguishing anomalies. Fig. \ref{fig: anomaly_score} shows the distributions of the anomaly scores over time of a participant's 24 anomalous and 20 non-anomalous feeding executions during leave-one-person-out cross validation. The blue and red shaded regions show the mean and standard deviation of non-anomalous and anomalous executions' anomaly scores, respectively. The score of non-anomalous executions shows a specific pattern of change with a smaller average and variance than that of anomalous executions, making anomalies easily distinguishable from non-anomalous situations.

The lower sub graphs of Fig. \ref{fig: reconstruction} also show the state-based threshold is capable of achieving a tighter anomaly decision boundary (red dash lines) than a fixed threshold over time. The expected anomaly scores (red curves) and the actual scores (blue curves) show a similar pattern of change. However, the expected score is lower than the actual score given an anomaly. Brown boxes show the time of anomaly detection where the first detection time matches with the initial increase of accumulated force.

We compared our LSTM-VAE with 5 other baseline methods through a leave-one-person-out cross-validation method. Given the training/testing dataset, we used data from 7 participants for training and tested with the data from the remaining 1 participant. Table \ref{table: eval_methods} shows our detector outperformed the other 5 baseline methods with higher area under the ROC curve (AUC) when using 4 hand-engineered features. The AUC is 0.044 higher than the next best method, HMM-GP. 

We also investigated the performance when using 17 sensory signals with the additional pre-training dataset. Our method resulted in the highest performance of AUC that is 0.064 higher than the next best method, EncDec-AD. It is also higher than the result from hand-engineered features. This indicates an LSTM-VAE is capable of modeling the heterogeneous high-dimensional multimodal signals and detecting anomalies among those signals without significant feature extraction effort. In this evaluation, we pre-trained each method using the pre-training dataset in addition to the dataset from the 7 participants. We then fine-tuned each method with the data from 7 participants. Note that we train an OSVM with the pre-traning dataset only and we did not succeed in training HMM-GP due to underflow errors resulting from the high-dimensional input.

\begin{figure}[t]
	\centering
	\includegraphics[trim=0.0cm 0.0cm 0.0cm 0.0cm, clip=true, width=6.5cm]{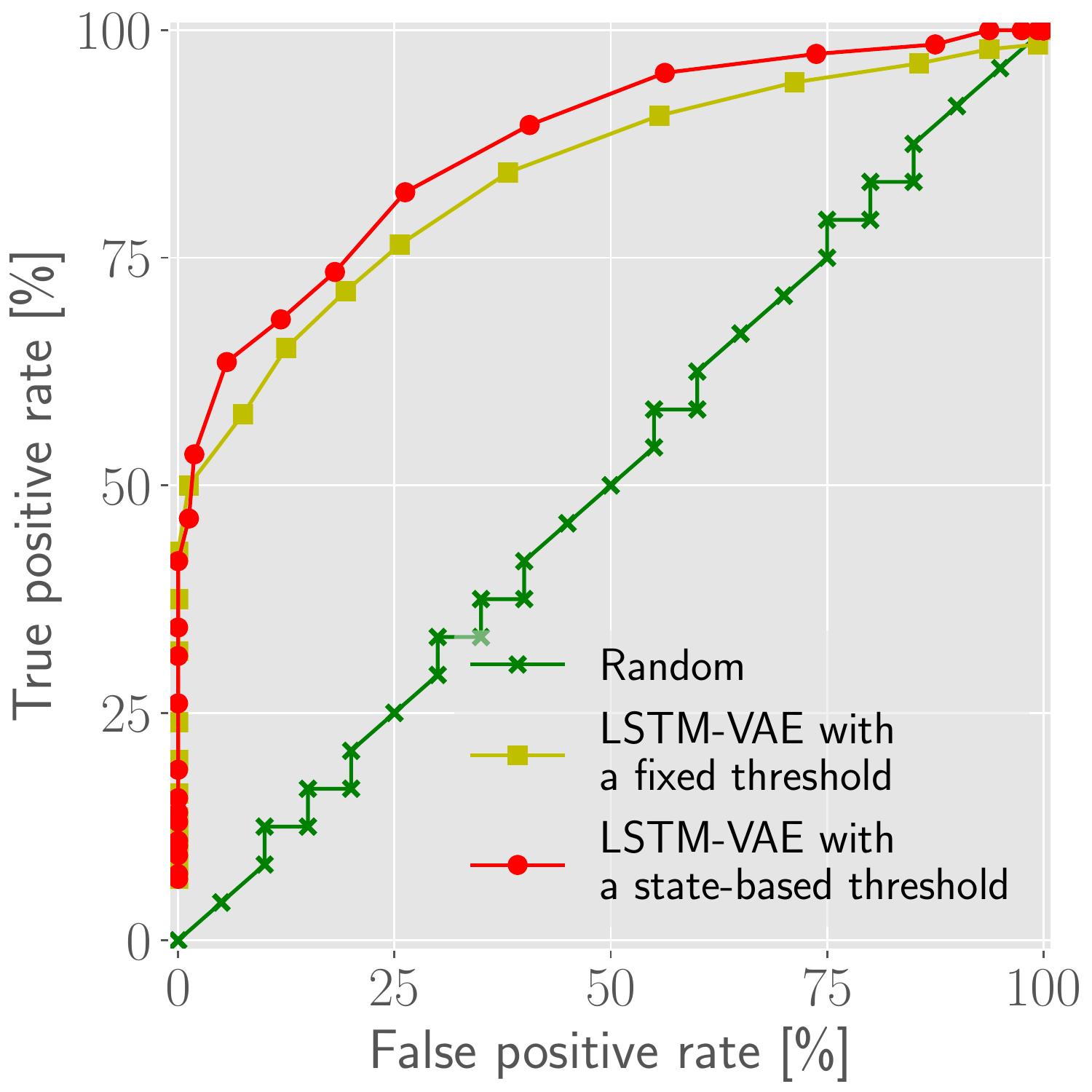} 
	\caption{\textit{Receiver operating characteristic (ROC) curves to compare the performance of LSTM-VAEs with and without a state-based threshold.}}
    \vspace{-1.5em}
	\label{fig: roc_thres}
\end{figure}

Fig. \ref{fig: roc_thres} shows ROC curves from an LSTM-VAE with two thresholding techniques. The red curve shows the result of the proposed state-based thresholding. The yellow curve shows the result of conventional fixed thresholding. The state-based thresholding resulted in higher true positive rates given the same false positive rates. In this evaluation, we used 17 sensory signals with the pre-training dataset.

\newcolumntype{T}{>{\centering\arraybackslash}m{3.7cm}<{}}
\newcolumntype{C}{>{\centering\arraybackslash}m{1.5cm}<{}}
\newcolumntype{M}{>{\centering\arraybackslash}m{1.5cm}<{}}
\begin{table*}[t] 
\caption{Comparison of the LSTM-VAE and 5 baseline methods with two types of inputs. Numbers represent the area under the ROC curve (AUC).}
\begin{center}
  \begin{tabular}{T|M|M|M|C|C|C}
    \hline    
    Input & Random & OSVM & HMM-GP & AE & EncDec-AD & LSTM-VAE \\ 
    \hline
    4 hand-engineered features & 0.5121 & 0.7427 & 0.8121 & 0.8123 & 0.7995 & \textbf{0.8564} \\     
    17 raw sensory signals & 0.5052 & 0.7376 & N/A & 0.8012 & 0.8075 & \textbf{0.8710} \\     
    \hline\noalign{\smallskip}
  \end{tabular}
\end{center}
\label{table: eval_methods}
\vspace{-1.5em}
\end{table*}

\section{Conclusion}
We introduced an LSTM-VAE-based anomaly detector for multimodal anomaly detection. An LSTM-VAE models the underlying distribution of multi-dimensional signals and reconstructed the signals with expected distribution information. The detector estimated the negative log-likelihood of multimodal input with respect to the distribution as anomaly score. By introducing a denoising autoencoding criterion and state-based thresholding, the detector successfully detected anomalies in robot-assisted feeding resulting in higher AUC than other 5 baseline methods in literature. Without significant effort of feature engineering, the detector with 17 raw inputs outperformed a detector trained with 4 hand-engineered features. Finally, we also showed the LSTM-VAE with the state-based decision boundary is beneficial for more sensitive anomaly detection with lower false alarms.

\section*{Acknowledgement}
\textit{This work was supported in part by NSF Award IIS-1150157, NIDILRR grant 90RE5016-01-00 via RERC TechSAge, and a Google Faculty Research Award. Dr. Kemp is a cofounder, a board member, an equity holder, and the CTO of Hello Robot, Inc., which is developing products related to this research. This research could affect his personal financial status. The terms of this arrangement have been reviewed and approved by Georgia Tech in accordance with its conflict of interest policies.}

\bibliographystyle{IEEEtran}
\bibliography{references}

\end{document}